\documentclass{article}

\usepackage[utf8]{inputenc}
\usepackage{amssymb,amsmath}
\usepackage{pifont}
\usepackage{graphicx,verbatimbox}
\usepackage{tcolorbox}

\usepackage{xspace}
\usepackage{booktabs}
\usepackage{multirow}
\usepackage{rotating}
\usepackage{color}
\usepackage{palatino}
\usepackage{multirow}
\usepackage[normalem]{ulem}
\usepackage{subfig}
\usepackage{colortbl}

\definecolor{citecolor}{RGB}{34, 149, 34}
\usepackage[pagebackref=true,breaklinks=true,letterpaper=true,colorlinks,linkcolor=black,citecolor=green,bookmarks=false]{hyperref}

\usepackage[parfill]{parskip}

\definecolor{Gray}{gray}{0.95}
\definecolor{DarkGray}{gray}{0.5}
\definecolor{LightCyan}{rgb}{0.88,1,1}
\definecolor{bisque}{rgb}{1.0, 0.89, 0.77}
\definecolor{blanchedalmond}{rgb}{1.0, 0.92, 0.8}
\definecolor{cosmiclatte}{rgb}{1.0, 0.97, 0.91}
\definecolor{cornsilk}{rgb}{1.0, 0.97, 0.86}

\title{\raisebox{-0.14in}{\includegraphics[width=0.08\linewidth]{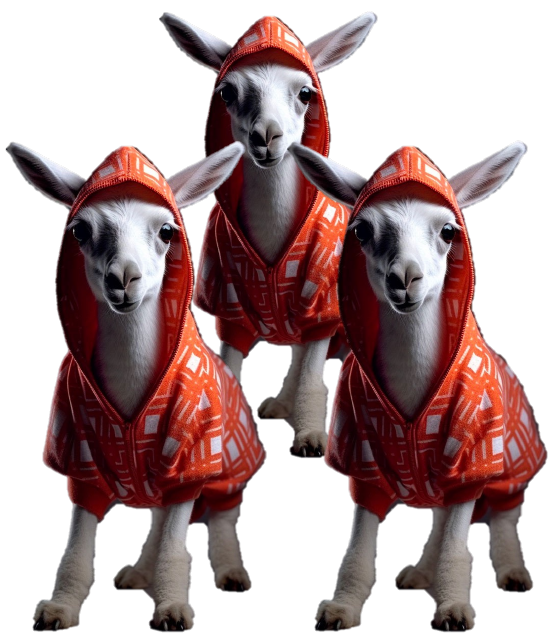}}{\bf SlimPajama-DC}: Understanding {\bf D}ata {\bf C}ombinations for LLM Training} 
\author{
 \large Zhiqiang Shen$^{\dagger}$ \hspace{0.2cm} Tianhua Tao$^{\dagger,\ddagger}$\hspace{0.2cm} Liqun Ma$^{\dagger}$\hspace{0.2cm} Willie Neiswanger$^{\S}$ \\ \hspace{0.2cm} Zhengzhong Liu$^{\dagger}$ \hspace{0.2cm} Hongyi Wang$^{\natural}$ \hspace{0.2cm} Bowen Tan$^{\natural}$ \hspace{0.2cm} Joel Hestness$^{\sharp}$ \\ \hspace{0.1cm} Natalia Vassilieva$^{\sharp}$ \hspace{0.1cm} Daria Soboleva$^{\sharp}$ \hspace{0.1cm} Eric  Xing$^{\dagger}$ \\[0.2cm] 
\scalebox{1.}{\hspace{-0.3cm} $^\dagger$MBZUAI\hspace{0.2cm} $^\ddagger$UIUC\hspace{0.2cm}$^\S$Stanford University\hspace{0.2cm}$^\natural$CMU\hspace{0.2cm} $^\sharp$Cerebras Systems}\\[1cm]
}

\date{~}

\begin{document}

\maketitle

\begin{abstract}
This paper aims to understand the impacts of various data combinations (e.g., web text, Wikipedia, GitHub, books) on the pretraining of large language models using SlimPajama. SlimPajama is a rigorously deduplicated, multi-source dataset, which has been refined and further deduplicated to 627B tokens from the extensive 1.2T token RedPajama dataset contributed by Together. We have termed our research as {\bf SlimPajama-DC}, an empirical analysis designed to uncover fundamental characteristics and best practices associated with employing SlimPajama in the training of large language models. During our research with SlimPajama, two pivotal observations emerged: {\bf (1)} Global deduplication vs. local deduplication. We analyze and discuss how global (across different sources of datasets) and local (within the single source of dataset) deduplications affect the performance of trained models. {\bf (2)} Proportions of highly-deduplicated multi-source datasets in the combination. To study this, we construct six configurations on SlimPajama dataset and train individual ones using 1.3B Cerebras-GPT model with Alibi and SwiGLU. Our best configuration outperforms the 1.3B model trained on RedPajama using the same number of training tokens by a significant margin. All our 1.3B models are trained on Cerebras 16$\times$ CS-2 cluster with a total of 80 PFLOP/s in bf16 mixed precision. We further extend our discoveries (such as {\em increasing data diversity is crucial after global deduplication}) on a 7B model with large batch-size training. Our SlimPajama-DC models are available at: \href{https://huggingface.co/MBZUAI-LLM/SlimPajama-DC}{link1} and the separate SlimPajama-DC datasets are available at: \href{https://huggingface.co/datasets/MBZUAI-LLM/SlimPajama-627B-DC}{link2}.
\end{abstract}

\tableofcontents

\section{Introduction}
\label{sec:introduction}

The success of modern large-scale models is deeply rooted in their training data. For large language models (LLMs), the emphasis is not merely on generic text but on ``diverse text''. To guarantee the model's linguistic expertise and its comprehensive understanding of the world, this text must span a broad spectrum of domains, genres, languages, and more. Consequently, the composition of the pretraining data domains, such as Github, Wikipedia, books, and web text like CommonCrawl, plays a critical role in the performance of large language models. In our research, we delve into the domain/source weightings of training data. Leveraging {\bf SlimPajama-DC}, we investigate two primary areas: (1) global-level and local-level deduplication, and (2) the efficacy of various combinations of thoroughly deduplicated datasets. The first emphasis basically encourages the model to be trained on all sources as no cross-domain overlaps inside, and the second helps us understand how to manage the integration and proportions of diverse domains, especially as datasets for LLM training continue to expand in variety.

\noindent{\textbf{Generic Deduplication.}} Multi-source datasets often combine data from various origins, each with its unique distribution of information. When training large language models, handling data redundancy is critical to ensure that the model generalizes well and does not exhibit undue biases, making training faster and more efficient. Highly deduplicated datasets ensure that the model isn't repeatedly exposed to the same or very similar data points, making the training more efficient. Redundant data can slow down convergence and might make the model overfit to frequently seen patterns. Deduplication helps in efficient utilization of the model's capacity. In general, deduplication is the process of removing duplicate data to address this redundancy. 

\noindent{\textbf{Global Deduplication {\em vs.} Local Deduplication.}} The global deduplication process removes duplicates from the entire combined datasets. When we're using data from multiple sources, there might be overlaps across sources. Global deduplication identifies and removes these overlapping instances irrespective of their source. In local deduplication, duplicates are removed within each individual source dataset before merging them. However, if two source datasets have overlapping data, those duplicates will still be present in the final combined dataset since deduplication was only done locally within each dataset. In most current open-source LLM training data~\cite{together2023redpajama,touvron2023llama,mosaicml23}, only local deduplication is performed within each data source, which neglects the redundancy across the different sources. Given the effects, global deduplication performed in SlimPajama is generally preferable for training large language models, especially when using multi-source datasets. It ensures a balanced representation of information and prevents the pitfalls associated with data redundancy. However, more hardware memory is naturally required by this strategy.

\noindent{\textbf{Different Combinations of Highly-deduplicated Datasets.}} A model trained on diverse data is more likely to generalize well across various tasks. It's exposed to a wider range of vocabulary, syntax, and semantics, enabling it to handle a broad scope of queries.  If diverse sources are chosen such that they represent different cultures, beliefs, and demographics, the model might be more balanced and less prone to biases. However, if many sources share common biases, the final dataset might amplify them. Different sources can provide both a breadth and depth of knowledge on various topics. Combining a technical dataset with a general news dataset, for example, would allow the model to understand both in-depth technical details and broad general knowledge.  It's crucial to note that data quality often outweighs the quantity. In this work, we aim to shed light on this fascinating perspective of comprehensive data combination on SlimPajama.

\noindent{\textbf{Specialization vs. Generalization Trade-off.}} In general, combining many specialized datasets can lead to a jack-of-all-trades model, which might not be as adept at specific tasks as a model trained on a specialized dataset. While the model can tackle a wide range of tasks, it might not have the depth of understanding that a specialized model might have for a particular domain. In this study, we also explore specialization and generalization ability using both individual and combined data sources.

The remainder of this paper is organized as follows. In Section~\ref{sec:dataset}, we elaborate the details of dataset statistics, token distributions, and data processing procedure. Section~\ref{sec:config} describes dataset combination configurations for this SlimPajama-DC study. Our model architecture and training details are provided in Section~\ref{arch_training_details}, followed by the results and analysis in Section~\ref{sec:experiments} on the range of various tasks in the zero- and few-shot settings. Section~\ref{sec:7B} presents an application of efficient Large Batch-size (LBS) training on a 7B model. Section~\ref{sec:related} reviews related work and Section~\ref{sec:conclusion} concludes this study.

\section{Dataset Overview}
\label{sec:dataset}

\subsection{Number of Tokens}
SlimPajama has a total of 627B tokens across different domains, as shown in Table~\ref{tab:data_proportion}. It includes validation and test sets with 500M tokens each, and these have been cleaned to ensure no overlap with the training data. For the {\bf SlimPajama-DC} study, our entire training dataset for each configuration contains 330B tokens after tokenization which is carefully selected from the original SlimPajama dataset. We tested different sampling strategies for different domains of our training data: (1) each token is trained only once during training, such as Commoncrawl, and (2) we perform more than one epoch for training on particular sources, such as the Wikipedia and Github domains. The detailed domain source proportions of various combinations are shown in Table~\ref{tab:six_conf}.

\begin{table}[h]
\centering
\resizebox{0.99\textwidth}{!}{
\begin{tabular}{l|cccccc}
\textbf{Dataset} & \multicolumn{1}{c}{SlimPaj.} & \multicolumn{1}{c}{RedPaj.} & \multicolumn{1}{c}{LLaMA-1}& \multicolumn{1}{c}{RefinedWeb} & \multicolumn{1}{c}{GPT3} &\multicolumn{1}{c}{MassiveText}   \\ \hline
Commoncrawl   & 52.2\%   & 72.6\%    & 67.0\% & 100\%   & 60.0\%  & 0.0\%   \\ 
 C4            & 26.7\%   & 14.4\%    & 15.0\% & 0.0\%   & 0.0\%   & 10.0\%  \\ 
GitHub        & 5.2\%    & 4.9\%     & 4.5\%  & 0.0\%   & 0.0\%   & 3.0\%  \\ 
 Books         & 4.2\%    & 2.1\%     & 4.5\%  & 0.0\%   & 16.0\%  & 27.0\%   \\ 
ArXiv         & 4.6\%    & 2.3\%     & 2.5\%  & 0.0\%   & 0.0\%   & 0.0\%   \\ 
 Wikipedia     & 3.8\%    & 2.0\%     & 4.5\%  & 0.0\%   & 3.0\%   & 2.0\%   \\ 
StackExchange & 3.3\%    & 1.7\%     & 2.0\%  & 0.0\%   & 0.0\%   & 0.0\%  \\ 
 WebText2      & 0.0\%    & 0.0\%     & 0.0\%  & 0.0\%   & 22.0\%  & 0.0\%    \\
MassiveWeb    & 0.0\%    & 0.0\%     & 0.0\%  & 0.0\%   & 0.0\%   & 48.0\%   \\
 News          & 0.0\%    & 0.0\%     & 0.0\%  & 0.0\%   & 0.0\%   & 10.0\%   \\  \hline
Total tokens  & 637B     &  1.2T     & 1.0/1.4T & 600B  & 300B    & 300B   \\ 
\end{tabular}
}
\vspace{-0.1in}
\caption{Data source proportions for various datasets.}
\label{tab:data_proportion}
\vspace{-0.1in}
\end{table}

\subsection{Dataset Token Frequency Statistics}

\begin{figure}[h]
\vspace{-0.25in}
\centering
    \subfloat[All Tokens] {
        \hspace{-0.2in}
        \includegraphics[width=0.36\linewidth]{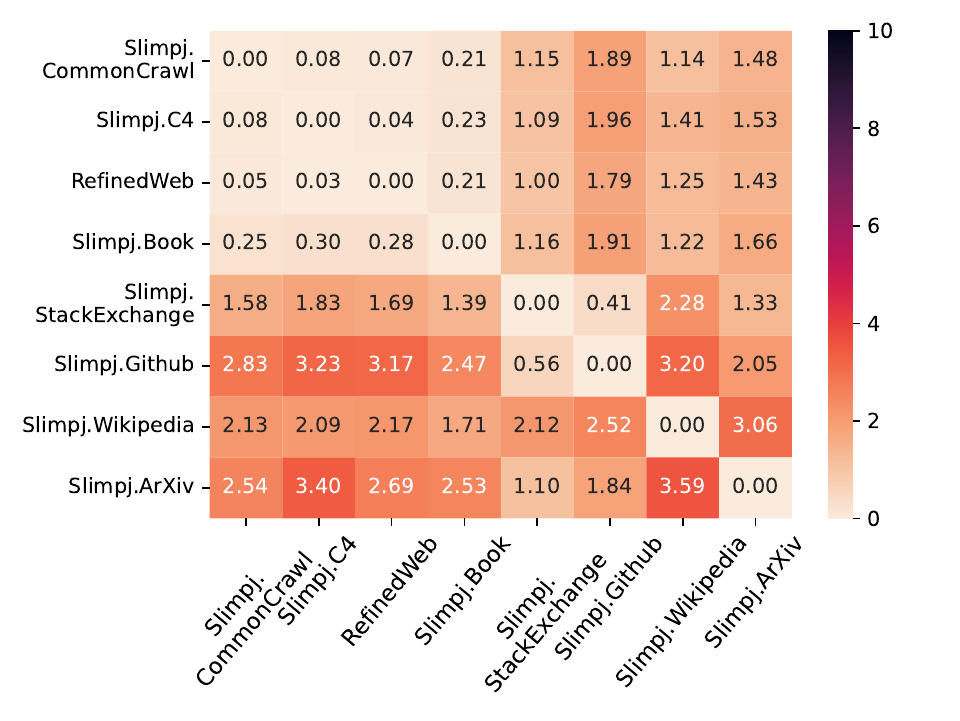}
        \label{fig:all_KL}
        \hspace{-0.1in}
    }
    \subfloat[Tokens Composed of Letters] {
        \includegraphics[width=0.33\linewidth]{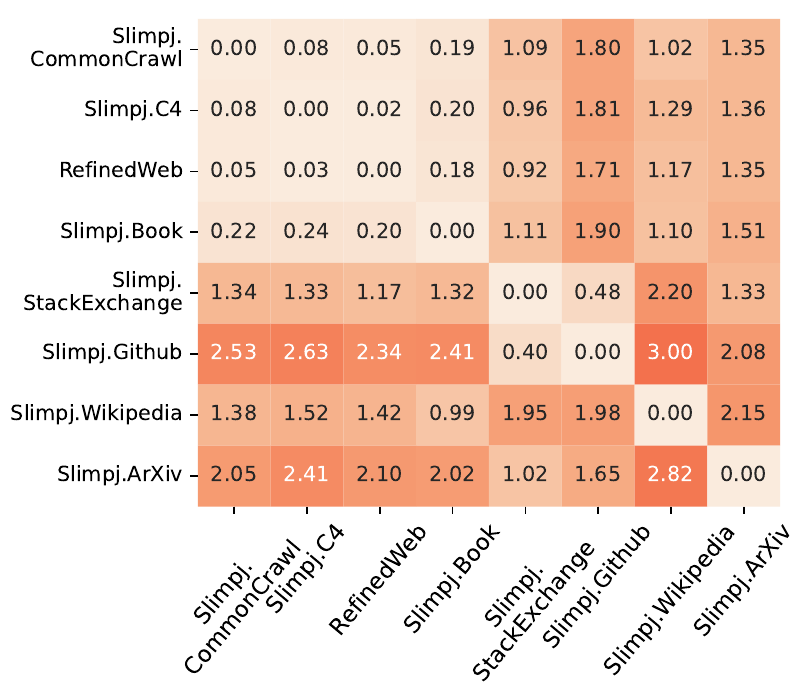}
        \label{fig:word_KL}
        \hspace{-0.1in}
    }
    \subfloat[Top 1000 Tokens] {
        \includegraphics[width=0.33\linewidth]{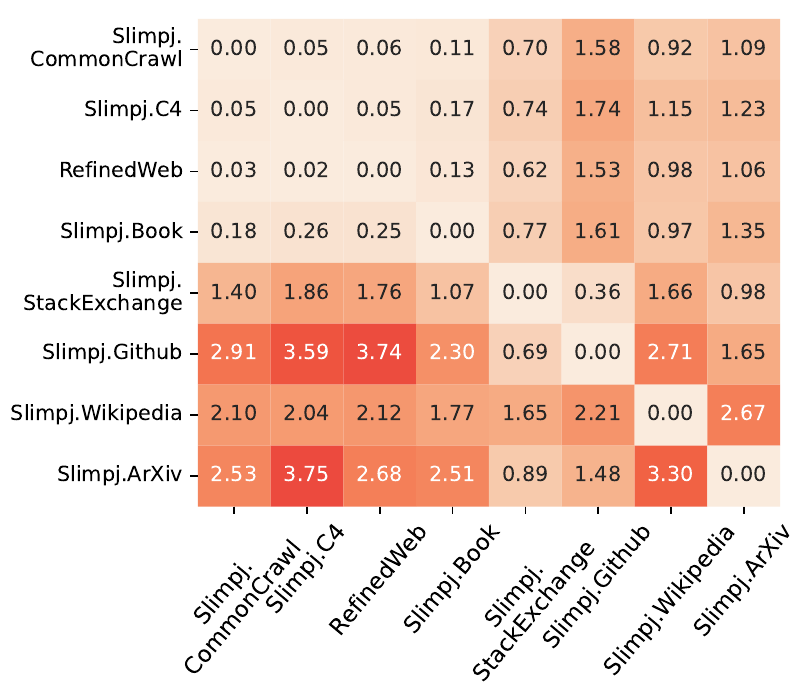}
        \label{fig:top1000_KL}
    }
    \\ \vspace{-0.15in}
    \subfloat[Numbers and Commonly Used Operators] {
        \hspace{-0.2in}
        \includegraphics[width=0.33\linewidth]{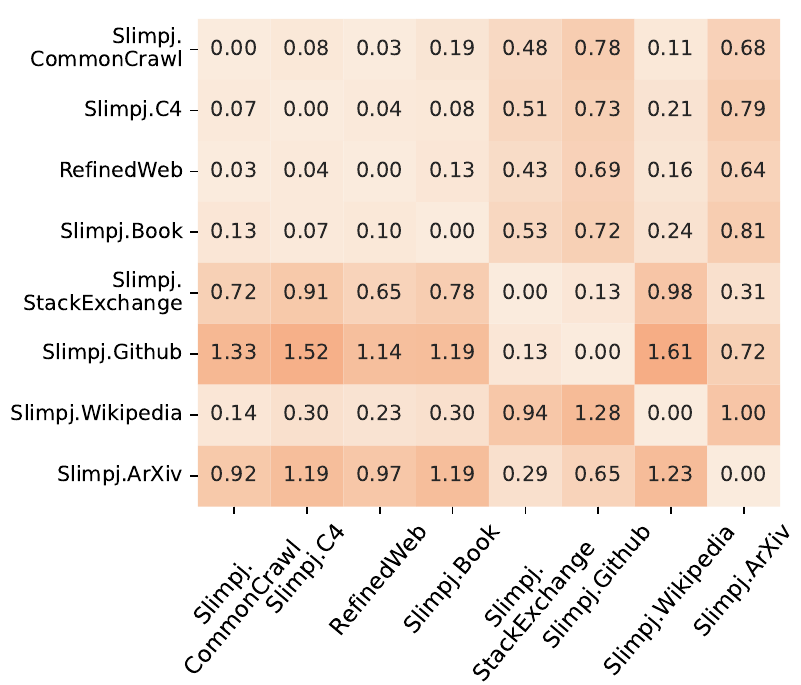}
        \label{fig:num_KL}
    }
    \subfloat[Whitespace Tokens] {
        \includegraphics[width=0.33\linewidth]{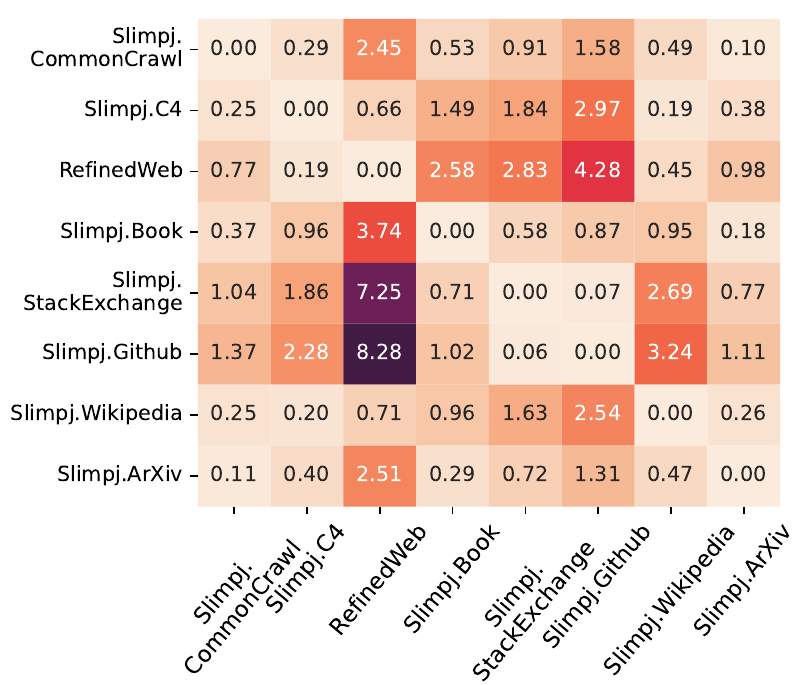}
        \label{fig:blank_KL}
    }
    \subfloat[Non-Alphanumeric Tokens] {
        \includegraphics[width=0.33\linewidth]{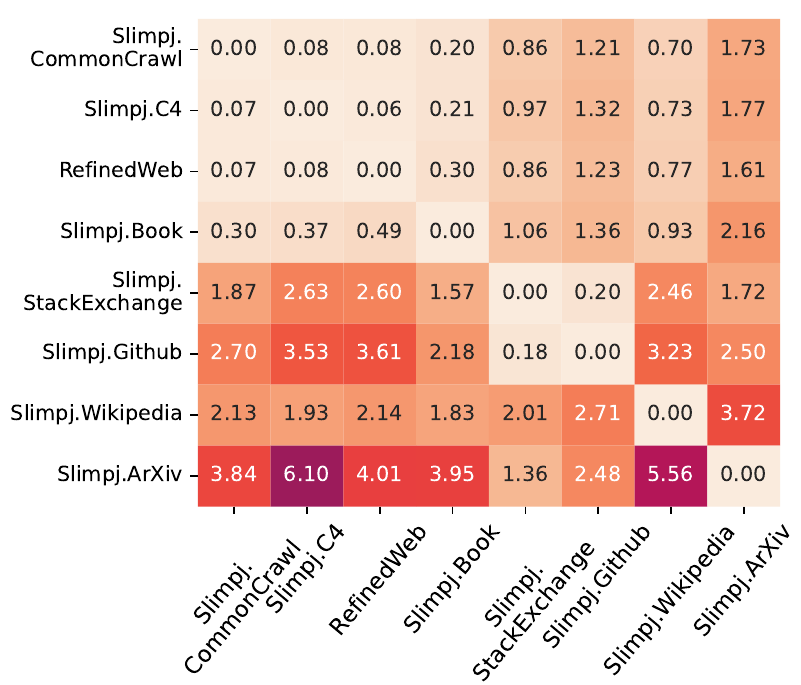}
        \label{fig:nonword_KL}
    }
\vspace{-0.1in}
\caption{Confusion matrix using KL divergence between the distributions of token statistics for different datasets.}
\vspace{-0.12in}
\end{figure}

To examine the similarity between various datasets in SlimPajama, we calculate the KL divergence between two domain distributions of token counts from different domain datasets, as shown in Fig. \ref{fig:all_KL}. 
Considering that different datasets often highlight varied types of tokens, such as GitHub prioritizing code and arXiv centering on academic content, we further investigate how these datasets vary in their distribution across subsets of tokens that exhibit unique features: 
(1) Tokens exclusively comprising letters (Fig. \ref {fig:word_KL}); 
(2) The union set of tokens with the top 1000 frequencies on each dataset (Fig. \ref {fig:top1000_KL});
(3) Numbers and commonly used operators, like `30', `+' and `=' (Fig. \ref {fig:num_KL}); 
(4) Whitespace Tokens, like `$\backslash$n$\backslash$n' and `$\backslash$t' (Fig. \ref {fig:blank_KL}); 
(5) Non-alphanumeric tokens, like `\#' and `====' (Fig. \ref {fig:nonword_KL}).

There exists a degree of similarity in the distribution of different token subsets among RefinedWeb, Book, C4, and CommonCrawl, as well as between Github and StackExchange. Notably, when it comes to the distribution of non-alphanumeric tokens, Arxiv differs significantly from most datasets. While on the distribution of whitespace tokens, Refinedweb shows notable distinctions in comparison to Github and StackExchange. Among numbers and commonly used operators, the distribution of all datasets is relatively consistent.

\subsection{Dataset Processing Procedure}

\begin{figure}[h]
\centering
\includegraphics[width=1.0\linewidth]{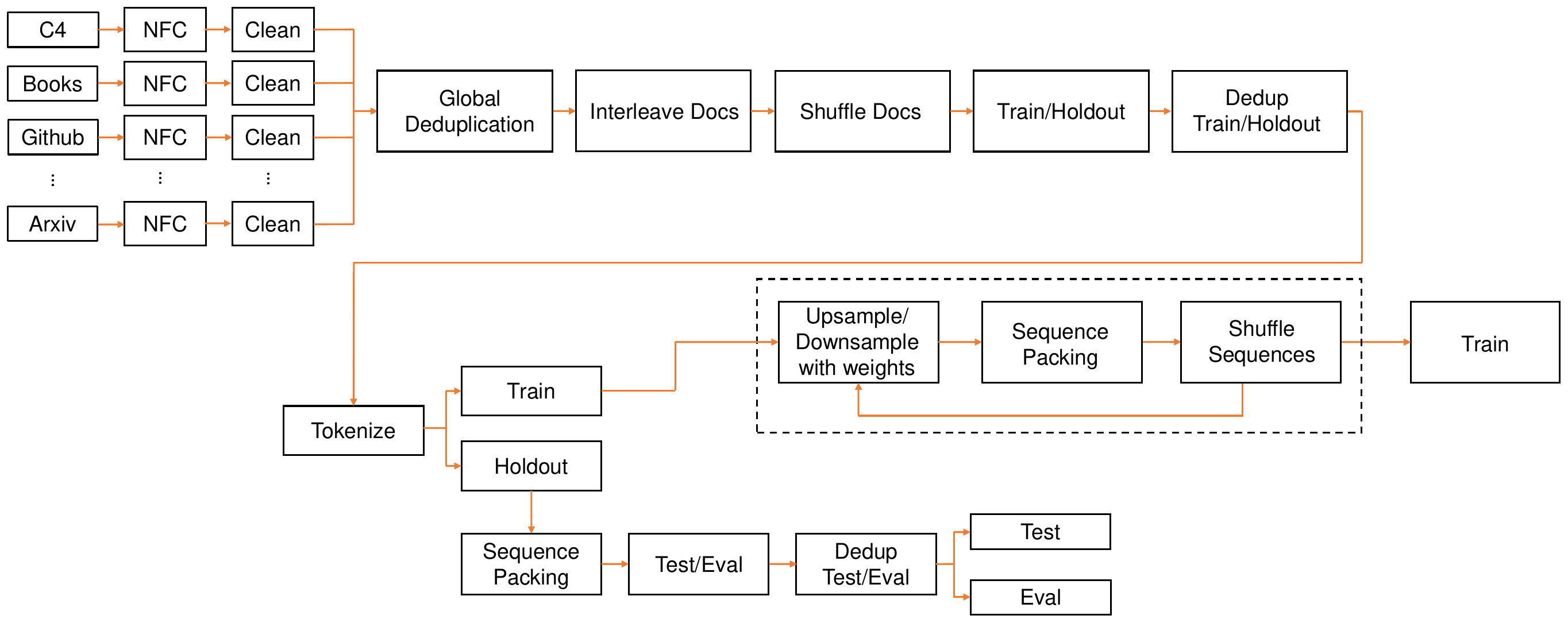}
\caption{SlimPajama preprocessing pipeline.}
\label{processing_pipeline}
\end{figure}

SlimPajama is constructed by filtering low-length documents and applying MinHashLSH deduplication to the 1.2T token RedPajama dataset to reduce it to 627B tokens. RefinedWeb~\cite{penedo2023refinedweb} shows that training on deduplicated data improves training compute efficiency and decreases the chance of LLMs generating memorized text from the dataset. By removing duplicate and low-length examples, it significantly improves the training compute efficiency and model performance. The overview of SlimPajama preprocessing pipeline is shown in Fig.~\ref{processing_pipeline} and the preprocessing code is available on \href{https://github.com/Cerebras/modelzoo/tree/main/src/cerebras/modelzoo/data_preparation/nlp/slimpajama}{GitHub}. 

\subsubsection{Low-length Document Filtering}

In the dataset processing procedure, an additional global filtering is performed to remove short, low-quality documents. After removing punctuation, consecutive spaces, newlines, tabs, and leading or trailing escape characters, documents with less than 200 characters are further filtered out. These documents typically contain only metadata and no useful information. A low-length filter is applied to every corpora other than Books and GitHub where it is found useful for short documents. The percentage of documents filtered out from each corpus within SlimPajama is detailed in Table~\ref{tab:filter_rates}. In total, this additional step removes 1.86\% of the documents.

\begin{table}[h]
\centering
\begin{tabular}{l|c|c}
 Data source   & Document filter rate & Byte duplication rate \\ \hline
Commoncrawl   & 0.02\%               & 63.76\%               \\ 
 C4            & 4.7\%                & 6.85\%                \\ 
GitHub        & 0.0\%                & 46.16\%               \\ 
 Books         & 0.0\%                & 2.01\%                \\ 
ArXiv         & 0.62\%               & 0.06\%                \\ 
 Wikipedia     & 0.0\%                & 2.24\%                \\ 
StackExchange & 0.32\%               & 0.20\%                \\ \hline
 Total         & 1.86\%               & 49.60\%               \\ 
\end{tabular}
\caption{Percentage of document low-length filter rates and data source byte duplication rates.}
\label{tab:filter_rates}
\end{table}

\subsubsection{Global Deduplication}

When building SlimPajama, it is observed that every corpus included in it contained duplicates with the most significant duplication found in CommonCrawl and GitHub. RefinedWeb~\cite{penedo2023refinedweb} also found similar rates of deduplication in the CommonCrawl data. It is most common to perform deduplication within each dataset source separately~\cite{touvron2023llama,together2023redpajama,zhang2022opt,gao2020pile} to reduce implementation complexity and meet resource constraints. This local deduplication approach does not have the ability to remove overlap between data sources which can be significant for web-scraped data. Instead, global deduplication removes duplication within and between each data source. Following~\cite{brown2020language,penedo2023refinedweb,biderman2023pythia,rae2021scaling}, global-level deduplication is performed using MinHashLSH algorithm. To facilitate global deduplication efforts and reproducibility for other researchers, a tool designed for scalable performance is offered under the above link. 

Specifically, global MinHashLSH deduplication is performed using a Jaccard similarity threshold of 0.8, document signatures constructed with preprocessed lowercase 13-grams, and schema following~\cite{leskovec2020mining}. To unify a representation of the same content, punctuation, consecutive spaces, newlines, tabs, and leading or trailing escape characters are removed. The level of deduplication performed per data source is presented in Table~\ref{tab:filter_rates}. The vanilla implementation of MinHashLSH did not scale to trillion token datasets like RedPajama without running out of memory. This is overcome by optimizing the memory usage and parallelization to perform deduplication on 64 CPU cores with 1.4TB peak memory, which can be easily decreased by creating multiple MinHashLSH objects to query.

\section{Dataset Combination Configurations}
\label{sec:config}

\subsection{SlimPajama}

\noindent{\bf Combination Strategies.} As shown in Table~\ref{tab:six_conf}, the adjusted domain weights establish a new training distribution. Using this distribution, we adopt a standard training approach to learn a consistent model architecture. This architecture remains unchanged across various domain weights and is trained using data from diverse combination distributions. Across different setups, we maintain the total training tokens to be the same. Our examination of domain weights in large language model training focuses on three main areas: 
1) Incrementally increasing the diversity of source combinations, as seen in configurations 1, 2, and 5, 6. 2) With consistent data sources, we explore varying domain proportions as in configurations 2 and 3. 3) We assess the significance of individual domain sources concerning the final model's performance as in configurations 3 and 4. Note that considering minimal impact of ArXiv and StackExchange, we have chosen to omit them in configuration 5 to preserve training resources and keep relatively sufficient training tokens for CommonCrawl. 4) Our final whole combination is applied in configuration 6 with the largest diversity.

\subsection{RefinedWeb}

 RefinedWeb~\cite{penedo2023refinedweb} is a massive English web dataset that is constructed using rigorous filtering and extensive deduplication of CommonCrawl. We use it as the comparison to our SlimPajama-DC CommonCrawl-only training.
 
\begin{table}[h]
\centering
\resizebox{0.99\textwidth}{!}{
\begin{tabular}{l|ccccccc|c}
 & Sub dataset & DC-1 & DC-2 & DC-3 & DC-4 & DC-5 & DC-6 &DC-7 \\ \hline
\multirow{7}{*}{SlimPajama}& Commoncrawl &  100.0\%   &  90.9\%    &  75.8\%  &  75.8\%  &  75.8\%  & 52.2\% & 0.0\% \\ 
&C4 &  0.0\%   &  0.0\%     &  0.0\%  & 0.0\%  &  0.0\% & 26.7\% & 0.0\% \\ 
&GitHub &  0.0\%   &   9.1\%    &  24.2\%  &  0.0\% &  9.1\%  & 5.2\% & 0.0\% \\ 
&Books &   0.0\%  &  0.0\%    & 0.0\%  & 0.0\% &  7.9\% & 4.2\%  & 0.0\% \\ 
&ArXiv &   0.0\%  &  0.0\%   & 0.0\%  & 0.0\%  &  0.0\%  & 4.6\%  & 0.0\% \\ 
&Wikipedia &  0.0\%   &  0.0\%    & 0.0\%  & 24.2\% &  7.3\%  &  3.8\% & 0.0\% \\ 
&StackExchange &  0.0\%   &  0.0\%   &  0.0\% & 0.0\%   & 0.0\% &  3.3\%  & 0.0\% \\  \hline 
RefinedWeb &  Commoncrawl   &  0.0\%   &0.0\%  & 0.0\%  &0.0\% & 0.0\% & 0.0\% & 100.0\% \\ \hline 
Total (Tokens)  &     & 330B  & 330B  & 330B  &  330B &  330B  &  330B  & 330B \\
\end{tabular}
}
\caption{Seven configurations of sub-dataset combinations in SlimPajama.}
\label{tab:six_conf}
\end{table}
\section{Network Architecture and Training Details}
\label{arch_training_details}

\subsection{Network Architecture}

\noindent{\bf Cerebras-GPT Architecture}~\cite{dey2023cerebras}.  Cerebras-GPT architecture shares similarities with those built on GPT-3~\cite{brown2020language}, particularly in the use of an autoregressive transformer decoder. However, a key difference lies in the attention mechanism employed. While GPT-3 utilizes a mix of dense and sparse-banded attention, Cerebras-GPT consistently uses dense attention across all decoder blocks. In terms of model dimensions, we either adhere to an aspect ratio of approximately 80 ($\text d_\text{model}$/$\text n_\text{layers}$) or maintain dimensions that are congruent with GPT-3 models. Additionally, all of our models are trained on a maximum sequence length of 2,048 tokens. The detailed architecture is shown in Table~\ref{tab:detailed_arch}.

\noindent{\bf Alibi}~\cite{press2021train}. Alibi introduces a more streamlined and efficient positional approach called {\em Attention with Linear Biases}. Rather than adding positional embeddings to word embeddings, ALiBi applies a bias to query-key attention scores, penalizing them based on their distance.

\noindent{\bf SwiGLU}~\cite{shazeer2020glu}. SwiGLU is an activation function which is a variant of GLU~\cite{dauphin2017language}. The formulation is as follows:
\begin{equation}
\operatorname{SwiGLU}(x, W, V, b, c, \beta)=\operatorname{Swish}_{\beta}(x W+b) \otimes(x V+c)
\end{equation}
where $x$  is a vector of the hidden representation at a particular position in the sequence. $W, V, b, c$ are the matrices and bias vectors, respectively.

\begin{table}[h]
\centering
\resizebox{1.0\textwidth}{!}{
\begin{tabular}{l|c|c|c|c|c|c|c}
 Model & n\_params & n\_layers & d\_model & n\_heads & d\_heads & batch size & learning rate \\ \hline
GPT-3 XL & 1.3B & 24 & 2,048 & 24 & 128 & 1M & 2.0$\times$10-4 \\
\bf Our DC   &    1.3B     &    24   &   2,048    &  24  &  128  &  2M  &  1.2$\times$10-2 \\
\end{tabular}
}
\caption{Detailed model sizes, architectures, and optimization hyper-parameters. Our LBS model details are presented in Appendix~\ref{sec:7B}.}
\label{tab:detailed_arch}
\end{table}

\subsection{Training Details}

\noindent{{\bf Tokenizer.}} We use an adapted GPT-NeoX~\cite{black2022gpt} BPE-based tokenizer similar to that used in GPT-2 for all our experiments, which has a vocabulary size of 50,277. Our entire training dataset for each configuration contains 330B tokens after tokenization, and each model takes about 2.5 days for training on Cerebras 16$\times$ CS-2S cluster.

\noindent{{\bf Optimizer.}} We employ the AdamW optimizer~\cite{loshchilov2017decoupled} to train our models, adopting these specific hyper-parameters: $\beta_1$ = 0.9, $\beta_2$ = 0.95, and eps = 1.0e-08. Our chosen learning rate follows a linear scheduler, culminating in a final learning rate that's 10\% of its peak value. Additionally, we apply a weight decay of 0.1, clip the gradient using a value of 1.0, and utilize a 150-step warmup.

\noindent{\bf Other Hyperparameters.} In our model, the filter size is 5,461, hidden size is 2,048 and attention dropout rate is 0. {\em SwiGLU} is used as the nonlinearity and {\em alibi} is used for position embedding. {\em Mixed precision} and {\em bfloat16} are employed during model training. More hyperparameters are shown in Table~\ref{tab:detailed_arch}.

\section{Results and Analysis}
\label{sec:experiments}

This section presents the analytical experiments and results on different combinations of SlimPajama. We first discuss the results following Huggingface Leaderboard Evaluation. Then, we demonstrate the importance of global deduplication and a diverse range of data sources in enhancing LLM's performance by conducting additional comprehensive evaluations across various topics. Finally, we visualize the training loss curves of different data domain combinations and provide insights on how they connect to the models' performance.

\subsection{Huggingface Leaderboard Evaluation with Harness}

Following the Huggingface Leaderboard Evaluation~\cite{open-llm-leaderboard}, we also assess our models on four key benchmarks using the Eleuther Language Model Evaluation Harness~\cite{eval-harness}. This unified framework facilitates the evaluation of generative language models across a broad scope of tasks. Specifically, our tests comprised: (1) AI2 Reasoning Challenge (25-shot)~\cite{clark2018think}; (2) HellaSwag (10-shot)~\cite{zellers2019hellaswag}; (3) MMLU (5-shot)~\cite{hendrycks2021measuring}; (4) TruthfulQA (0-shot)~\cite{lin2022truthfulqa}. 
As shown in Table~\ref{tab:six_conf_res}, with the exception of DC-4, our average results are all better than RedPajama-1.3B which is also trained on 330B tokens. Among our combinations, the DC-1 (which relies solely on SlimPajama Commoncrawl) achieves the highest scores for ARC and MMLU among all tested configurations. Yet, its performance on TruthfulQA ranks at the bottom. On the other hand, DC-6 obtains the top average accuracy across all SlimPajama data combinations, as well as standing out with the best results on HellaSwag. From DC-1, 2, 5, and 6, it is clear that more domain combinations with diverse training data bring better overall accuracy. Moreover, a potential strategy to harness the strengths of each configuration might involve a sequential training process on DC-1, DC-6, and DC-7.

Furthermore, SlimPajama uses global deduplication across all sources. This suggests that merging all domains typically yields better results than selective combinations as the elimination of overlaps among different datasets. This also highlights the importance of global deduplication and a diverse range of data sources for LLM overall performance.

\begin{table}[t]
\centering
\resizebox{1.0\textwidth}{!}{
\begin{tabular}{l|c|ccccc}
 \textbf{Model} &  Training Tokens &  \textbf{Average} & \textbf{ARC} & \textbf{HellaSwag} & \textbf{MMLU} & \textbf{TruthfulQA} \\ \hline
   Cerebras-GPT-1.3B~\cite{dey2023cerebras}  & 26.3B  & 33.5   &     26.3    &   38.5       &     26.6   &   42.7 \\ 
   OPT-1.3B~\cite{zhang2022opt}     & 180B  &  36.6  &  29.2   &  54.4  &   24.4     &      38.5         \\
   Pythia-1.4B~\cite{biderman2023pythia}  & 300B  & 37.3  &  31.9  & 52.8   &  25.6  &  38.8    \\
   GPT-neo-1.3B~\cite{gpt-neo}   &  300B &  36.0      &   31.2   &   48.5    &   24.8  &   39.6 \\ 
   RedPajama-1.3B~\cite{together2023redpajama}   & 330B   &   38.0   &    37.2 &    55.8   &      24.9  &   34.3       \\ 
   TinyLlama-1.1B~\cite{zhang2024tinyllama}    & 3T &  39.4 &  33.6  & 60.3 &  25.9  &   37.6         \\    
   Olmo-1.2B~\cite{groeneveld2024olmo}  & 3T & 39.4 & 34.4 & 	63.7  &  26.4 &  33.0 \\  \hline
   Our DC-1-1.3B            & 330B & 38.5    &    36.3  &     56.0     &     27.0      &      34.8       \\ 
   Our DC-2-1.3B            & 330B &  38.4    &   33.9    &     55.5     &     25.7     &     38.6    \\ 
   Our DC-3-1.3B            & 330B &  38.5     &   35.2   &    54.7     &   25.7  &     38.3         \\ 
   Our DC-4-1.3B            & 330B & 37.6     &  33.4  &   53.3     &   26.0    &   37.6         \\  
   Our DC-5-1.3B            & 330B &   38.6    &   34.7     &     56.0     &  25.6    &   38.0         \\ 
   Our DC-6-1.3B            & 330B & \bf 40.0 &   33.7  & 61.0 &  26.9 & 38.4  \\\hline
   Our DC-7-1.3B$^{\ddag}$  & 330B &   \bf 41.0     &  35.1    &      64.7     &    26.2     &    37.9   \\ 
\end{tabular}
}
\vspace{-0.1in}
\caption{Results of seven dataset combination configurations following Huggingface Leaderboard~\cite{open-llm-leaderboard} using Harness~\cite{eval-harness}. $^{\ddag}$ is the RefinedWeb CC.}
\label{tab:six_conf_res}
\vspace{-0.1in}
\end{table}

\subsection{More Evaluations}

As in Table~\ref{tab:six_conf_more_res}, we present additional evaluations across various domains to investigate the fine-grained capabilities offered by different data combinations. Except for DC-7 (trained on RefinedWeb dataset), incorporating more sources, such as DC-1, 2, 5 and 6, typically leads to improved average performance. 

Upon analysis, we find that specific mixtures perform well in particular evaluation benchmarks. For example, DC-1 obtains the highest accuracy in the ARC challenge and Race. Meanwhile, DC-4 outperforms others in the Winogrande and Xstory cloze, and DC-5 emerges as the top performance in the WSC273 evaluation. Moreover, all of our configurations except for DC-4 are superior in the average performance over the comparisons of GPT-neo-1.3B~\cite{gpt-neo} and RedPajama-1.3B~\cite{together2023redpajama}.

\noindent{\bf Risk of random guessing score on 1.3B models.} 
It is widely recognized that small models, such as the 1.3B variant, may struggle to achieve satisfactory predictions on specific benchmarks like MMLU. Their results could resemble random choices, not truly capturing the model's actual capabilities. To more accurately showcase a model's true potential and reflect the ability of different data combinations, we introduce a novel metric $\text{RRGS}$ (risk of random guessing score) to evaluate the degree of random guessing. Since 25\% in MMLU represents the baseline score for a guess, this metric evaluates the variance using average $\ell_1$ distance around this base value across all sub-items. A larger variance would suggest a reduced likelihood of predictions resulting from mere chance. Given a MMLU score vector $X$ of length $N$ with sub-item scores $s_1, s_2, \dots, s_n$, $\text{RRGS}$ can be formulated as:
\begin{equation}
\text{RRGS}=1-\frac{1}{N}\sum_{i=1}^{N}(|s_i- 0.25|)
\end{equation}
where $i$ is the index of sub-item in MMLU and $N$ is the number of items of MMLU. This metric utilizes the probabilities of variance to baseline 25\%, aiming to assess the extent to which a model's prediction resembles random guessing on the MMLU benchmark. 
The metric has three variations: (1) Consider only items with scores exceeding 25\%, i.e., $i\in{\text{\{positive item set\}}}$. (2) Focus solely on items with scores less than 25\%, i.e., $i\in{\text{\{negative item set\}}}$. (3) Include all items and sum them up.
The results are shown in Table~\ref{tab:random_guessing}. Generally, a model with a higher MMLU average score will have a low risk of random guessing probability.

It is also crucial to employ a broader and more diverse set of benchmarks, such as in Table~\ref{tab:six_conf_more_res}. Additionally, for a detailed understanding, we have cataloged the complete MMLU results for every sub-item in Table~\ref{tab:mmluapp}. This offers a lens into the knowledge assimilated by the pretrained models within each sub-domain on this comprehensive benchmark.

\begin{table}[t]
\centering
\resizebox{0.99\textwidth}{!}{
\begin{tabular}{l|cc|cccccc|c|c}
 \textbf{Eval} & \textbf{Neo} & \textbf{RedPaj.}  & \textbf{DC-1} & \textbf{DC-2}  & \textbf{DC-3} & \textbf{DC-4} &  \textbf{DC-5} & \textbf{DC-6} & \textbf{DC-7} & \textbf{LBS}  \\ 
 &  \multicolumn{2}{c|}{\bf 1.3B}  &   \multicolumn{7}{c}{\bf 1.3B}   &  \bf 7B  \\ \hline
Humaneval (p@1)   & - &   -  & -  &   - & - & - &  - & - & - & 9.5 \\ \hline
ARC\_easy        & 61.1 & 66.7   & 66.1	    &	66.9 		&	66.4 	&	65.5 &	66.5 & \uline{\textbf{67.8}}	&	66.8 	&  74.7     \\
ARC\_challenge   & 25.9 &  33.5  & \uline{\textbf{36.3}} 	&	33.9	&	35.2 	&	33.4 &	34.7 & 33.7	&	35.1 	&  44.3  \\
Boolq            & 62.0 & 55.6   & 63.4 	&	\uline{\textbf{65.6}}	&	64.2 	&	50.6 &   62.5 & 59.6	&	61.7 	&  66.9     \\
PIQA             & 71.1 &  72.4  & 70.8     &	69.2   &	68.6 	&	67.8  &	70.7 & \uline{\textbf{72.9}}		&  75.7 	&  77.4     \\
Race             & 34.1 & 34.4   & \uline{\textbf{37.3}} 	&	36.7 	&	36.5 	&	34.6  &	\uline{\textbf{37.3}} & 36.8	&	36.6 	&  38.2  \\
Winogrande       & 54.9 & 60.5   & 60.3 	&	59.7 	&	60.1 	&	\uline{\textbf{60.5}} &	59.8 & 59.4	&	61.2   &  64.4   \\
Openbookqa       & 33.6 &  33.0  & 35.6	    &	34.8	&	34.0	&	34.4 &	34.0 & \uline{\textbf{36.0}}	&	37.4    &  39.8   \\
COPA             & 69.0 & 77.0   & 70.0	    &	73.0	&	74.0	&	70.0	&	75.0 & \uline{\textbf{81.0}}	&	81.0    &  86.0  \\
WSC273           & 75.1 &  78.0  & 76.2 	&	78.0 	&	76.9 	&   76.6 	&	\uline{\textbf{81.0}} & 76.2	&	79.5    &  85.0   \\
Swag             & 67.8 & 68.8   & 69.2 	&	68.5 	&	67.8 	&	68.3 	&	70.1 & \uline{\textbf{70.2}}	&	70.0    &  73.8   \\
Pawsx*           & 50.6 & 51.5   & 51.4 	&	52.3 	&	52.2 	&	50.5 	&	53.1 & \uline{\textbf{55.2}}  &	50.8    &  54.7    \\
Xstory\_cloze*   & 51.1 & 51.5   & 51.0	    &	51.3	&	51.5	&	\uline{\textbf{52.2}} &	52.0 & \uline{\textbf{52.2}}		&	51.6    &  55.3  \\ \hline
Average          & 54.7 & 56.9   & 57.3     &   57.5    &   57.3  & 55.4 &   58.1 &  \uline{\textbf{58.4}}&  58.9 &  63.4  \\
\end{tabular}
}
\vspace{-0.1in}
\caption{Results of seven dataset combination configurations of 1.3B models and our LBS-7B model. Details of LBS-7B are presented in Appendix~\ref{sec:7B}. 
ARC\_easy and ARC\_challenge are evaluated using 25-shot. All other evaluation benchmarks are tested on 0-shot. * represents the results are averaged across multiple sub-items inside each benchmark dataset.}
\label{tab:six_conf_more_res}
\end{table}

\begin{table}[h]
\centering
\resizebox{0.73\textwidth}{!}{
\begin{tabular}{l|cccccc|c}
 &  DC-1 & DC-2 & DC-3 & DC-4 & DC-5 & DC-6 & DC-7 \\ \hline
 MMLU                       &   0.27 &  0.257    & 0.257 & 0.260 &  0.256 & 0.269 & 0.262 \\ \hline
 $\text{RRGS}_\text{pos}$   &  0.964 & 0.964  & 0.965 & 0.970 &  0.968 &\bf 0.954 & 0.963\\  
$\text{RRGS}_\text{neg}$    & 0.974 & 0.973  & 0.974 &\bf 0.969  &  0.975 & 0.972 & 0.973\\
 $\text{RRGS}_\text{all}$    & 0.968 & 0.968  & 0.969 & 0.970 &  0.971 & \bf 0.962 & 0.967\\
\end{tabular}
}
\vspace{-0.1in}
\caption{Evlauation of random guessing probability on sub-items of MMLU. }
\label{tab:random_guessing}
\vspace{-0.1in}
\end{table}

\subsection{Training Loss}

\begin{figure}[h]
\vspace{-0.15in}
\centering
\includegraphics[width=0.79\linewidth]{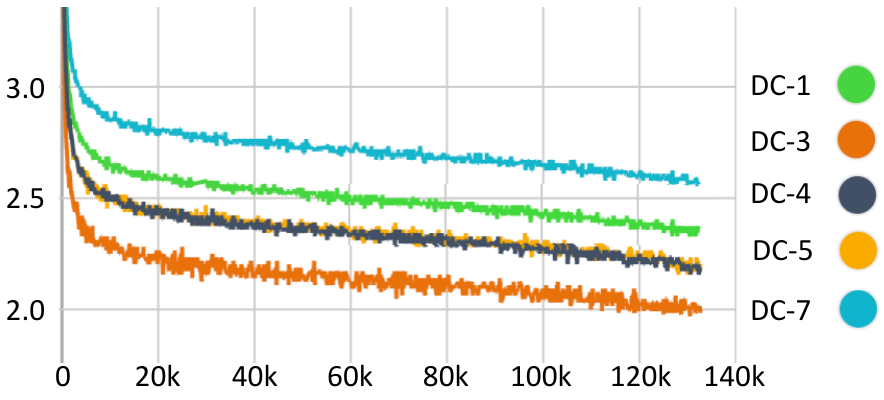}
\vspace{-0.15in}
\caption{Illustration of training loss curves. DC-2's curve closely resembles those of DC-4 and 5, so it has been excluded from the figure for clarity.}
\label{loss_curves}
\vspace{-0.06in}
\end{figure}

Fig.~\ref{loss_curves} presents the training loss curves for various data combinations, from which several insights can be observed: 1) While DC-7 demonstrated the highest average accuracy in our quantitative evaluations, its training loss was also the most substantial. This suggests that a lower training loss doesn't necessarily correlate directly with superior model performance. 2) DC-3, with a considerable portion of its data coming from code domain, exhibited the lowest training loss. This implies that as the amount of code in training increases, the training loss diminishes. 3) The training loss values for other combinations appeared to be relatively consistent with one another.

\section{Related Work}
\label{sec:related}

\subsection{RedPajama, SlimPajama and Others.} 

 RedPajama~\cite{together2023redpajama} aims to develop open-source large language models and begins by replicating the LLaMA training dataset~\cite{touvron2023llama}, which boasts over 1.2 trillion tokens. This collaborative effort involves entities such as Together, Ontocord.ai, ETH DS3Lab, Stanford CRFM, Hazy Research, and the MILA Québec AI Institute. SlimPajama~\cite{cerebras2023slimpajama} stands as the highly deduplicated, multi-source, open-source dataset tailored for training large language models. This dataset emerged by refining and eliminating duplicates from the whole 1.2T token RedPajama dataset. Through meticulous filtering of subpar data and repetitive content, it reduced the dataset size by 49.6\%, scaling it down from 1.2T to 627B tokens. SlimPajama provides superior quality and computational efficiency for training tasks than the original RedPajama dataset. Other efforts also have been made in this direction to construct diverse datasets, such as Pile~\cite{gao2020pile}. It is an English text corpus of 825 GiB, which is designed for the training of large-scale language models with increased training dataset diversity to improve general cross-domain knowledge and downstream generalization capability. It contains a combination of 22 distinct, high-quality subsets. These subsets incorporate both pre-existing and freshly curated data, with a significant portion sourced from scholarly or professional domains.

\subsection{Data Processing and Optimization Approaches}

There have been several advancements in data processing and optimization. The seminal method of importance sampling~\cite{kloek1978bayesian} stands out as a Monte Carlo approach designed to evaluate attributes of a particular distribution, even when the samples are drawn from a distribution that differs from the one under exploration. SlimPajama's deduplication mechanism is an adaptation of importance sampling, incorporating a heuristic that values unique data points. Recently, several data selection frameworks~\cite{katharopoulos2018not,gururangan2020don,sorscher2022beyond,xie2023data} have been introduced, inspired by the concept of importance sampling. Among them, DSIR~\cite{xie2023data} presents a framework for the data selection challenge by aiming to choose a subset from a large, unlabeled raw dataset that aligns with a specific target distribution, given a set of unlabeled target examples. It builds upon the traditional importance resampling method, adapting it for data selection in large-scale models. DSIR operates as a scalable algorithm, determining importance weights within a reduced feature space and then selecting data based on these importance resampling weights.
In~\cite{sorscher2022beyond}, it delves into the relationship between error scaling and dataset size. Its theoretical exploration suggests that by using a robust data pruning metric, which prioritizes which training examples to remove, the proposed method can suppress traditional power law scaling, potentially reaching exponential scaling for pruned dataset sizes.

\subsection{Data Combination for Training Large Language Models}

The training of large language models, such as GPT~\cite{radford2018improving,radford2019language,brown2020language} and BERT~\cite{devlin2019bert}, requires significant amounts of data to capture and generalize over the vast intricacies of human language. As a result, researchers often combine data from various sources, such as web text, Github, Books, ArXiv, Wikipedia, etc. There are some related work and difficulties that have been explored in the context of data combination for training large language models. (1) Concatenation of diverse datasets: One of the simplest methods for combining data is to concatenate various corpora, covering diverse topics, styles, and sources. This ensures that the model gets a broad view of the language. (2) WebText and similar corpora: For OpenAI’s GPT-2, a dataset called WebText~\cite{radford2019language} was curated by scraping content from the internet. This kind of data provides a rich mix of formal, informal, factual, and opinionated text, thus offering diverse training material. (3) Balancing and weighting: Simply combining data may lead to issues if one source is overrepresented. Prior studies have applied weights to different data portions or ensure that the combined dataset is balanced in terms of sources, styles, and other criteria. For instance, DoReMi~\cite{xie2023doremi} first trains a small proxy model using group distributionally robust optimization across domains, generating domain weights (or mixture proportions) without relying on information from subsequent tasks. Following this, they utilize these domain weights to resample a dataset, on which then train a full-size model. (4) Multimodal Training: Combining text with other data forms, like images or sounds, can also enhance language model training, especially for tasks that require understanding across modalities.

\subsection{Large Batch Training for Large Language Models}
Large language models inherently possess a structure that supports parallelization, especially when optimized using techniques that allow for batch training. When computational resources permit, large batch sizes are favored to expedite the training of large models containing potentially millions or billions of parameters. At a fundamental level, larger batch sizes enhance the quality of each gradient update since they consider a more considerable chunk of the dataset. Conversely, a smaller batch size means that model parameter updates are based on gradients derived from a limited dataset portion. This smaller dataset slice might not comprehensively capture the intricate relationships between features and labels. Therefore, it might seem that larger batch sizes consistently offer advantages in training. 
However, ~\cite{keskar2016large} pointed out that this perspective does not factor in the model's capacity to generalize to new, unseen data, nor the intricate, non-convex optimization landscape of contemporary large models. In practice, multiple studies~\cite{hoffer2017train,keskar2016large} have demonstrated that while larger batch sizes might hasten convergence, they can impair a model's generalization to new datasets, irrespective of the deep network type. This observed disparity has been named as the {\em Generalization Gap}. A method~\cite{hoffer2017train} to address this gap involves starting from a smaller batch size and gradually enlarging it as training advances. In our study, we explore this problem through a new and unique angle of progressive weight decay training.

\section{Conclusion}
\label{sec:conclusion}

We have presented {\bf SlimPajama-DC}, a comprehensive study on understanding the data domain weights and combinations for training large language models. Notably, SlimPajama-DC can operate on compact models, and its advantages can be seamlessly transferred to models that are several times larger. This leads to a remarkable acceleration in training on the SlimPajama with the optimal sampling probabilities across domains for larger models. Through this, we aim to spark further exploration into data-centric methods to enhance the understanding and efficiency of large language model pretraining.

{\small
\bibliographystyle{ieee_fullname}
\bibliography{my}
}

\newpage

\appendix

\section*{\LARGE Appendix}

\section{Data Proportion Details}

\begin{table}[h]
\centering
\resizebox{0.84\textwidth}{!}{
\begin{tabular}{l|cc|cc|cc}
 \textbf{Dataset} & \multicolumn{2}{c|}{Slimpajama} & \multicolumn{2}{c|}{Redpajama} & \multicolumn{2}{c}{LLaMA 1}  \\ \hline
Commoncrawl   & 52.2\%  & 333B & 72.6\%  & 878B & 67.0\% & 670/938B \\ 
 C4            & 26.7\%  & 170B & 14.4\%  & 175B & 15.0\% & 150/210B \\ 
GitHub        & 5.2\%   & 33B  & 4.9\%   & 59B  & 4.5\%  & 45/63B   \\ 
 Books         & 4.2\%   & 27B  & 2.1\%   & 26B  & 4.5\%  & 45/63B   \\ 
ArXiv         & 4.6\%   & 29B  & 2.3\%   & 28B  & 2.5\%  & 25/35B   \\ 
 Wikipedia     & 3.8\%   & 24B  & 2.0\%   & 24B  & 4.5\%  & 45/63B   \\ 
StackExchange & 3.3\%   & 21B  & 1.7\%   & 20B  & 2.0\%  & 20/28B  \\ 
 Total         & 100.0\% & 637B & 100.0\% & 1.2T & 100\%  & 1.0/1.4T \\ \hline
             & \multicolumn{2}{c|}{RefinedWeb} & \multicolumn{2}{c|}{GPT3} &\multicolumn{2}{c}{MassiveText}  \\ \hline
Commoncrawl   & 100\%   & 600B & 60.0\%  & 180B & 0.0\%  & 0\\ 
 C4            & 0.0\%   & 0B   & 0.0\%   & 0    & 10.0\% & 30B\\ 
GitHub        & 0.0\%   & 0B   & 0.0\%   & 0    & 3.0\%  & 9B\\ 
 Books         & 0.0\%   & 0B   & 16.0\%  & 48B  & 27.0\% & 81B\\ 
Wikipedia     & 0.0\%   & 0B   & 3.0\%   & 9B   & 2.0\%  & 6B\\ 
WebText2      & 0.0\%   & 0B   & 22.0\%  & 66B  & 0.0\%  & 0 \\ 
MassiveWeb    & 0.0\%   & 0B   & 0.0\%   & 0    & 48.0\% & 144B \\
 News          & 0.0\%   & 0B   & 0.0\%   & 0    & 10.0\% & 30B \\  \hline
Total         & 100.0\% & 600B & 100.0\% & 300B & 100.0\% & 300B  \\ 

\end{tabular}
}
\caption{Detailed data source proportions for various datasets.}

\label{tab:appendix_data}
\end{table}

\section{MMLU}

In this section, we provide the detailed item-by-item results in MMLU, as shown in Table~\ref{tab:mmluapp}, it is interesting to notice that on some sub-domains in MMLU, the results from our configured 1.3B models are even better than   GPT-3 175B and LLaMA2 7B models.

\begin{table*}
  \hspace{-0.2in}
  \setlength{\tabcolsep}{4pt}
   \scalebox{0.68}{
  \begin{tabular}{lrcccccccc}
    \toprule
    &  & \multirow{2}{*}{\textbf{GPT-3}} & \multirow{2}{*}{\textbf{Llama2}} & \multicolumn{6}{c}{\textbf{SlimPajama-DC 1.3B}}    \\
    \cmidrule(lr){5-10} 
    &  &             \bf 175B            &            \bf  7B           & DC-1 & DC-2 & DC-3 & DC-4 & DC-5 & DC-6  \\
\midrule
Abstract Algebra & STEM & \textbf{30.0} & 29.0 & 27.0 & 26.0           & 25.0 & 27.0 & 28.0 & 23.0 \\ 
  Anatomy & STEM & \textbf{48.0} & 37.0 & 23.0 & 23.0                  & 27.4 & 34.1 & 25.9 & 28.1 \\ 
Astronomy & STEM & \textbf{49.0} & 33.6 & 25.0 & 19.7                  & 23.0 & 27.0 & 21.7 & 23.7 \\ 
Business Ethics & Other & \textbf{46.0} & 40.0 & 24.0 & 22.0           & 26.0 & 24.0 & 30.0 & 25.0 \\ 
Clinical Knowledge &Other&\textbf{48.0} & 35.1 & 30.2 & 26.8           & 24.9 & 18.9 & 25.7 & 24.2 \\ 
  College Biology & STEM & \textbf{45.0} & 37.5 & 23.6 & 24.3          & 27.1 & 25.7 & 23.6 & 22.9 \\ 
College Chemistry & STEM & 26.0 & \textbf{32.0} & 26.0 & 19.0          & 29.0 & 19.0 & 21.0 & 21.0 \\ 
  College Computer Science & STEM & \textbf{46.0} & 29.0 & 37.0 &36.0  & 32.0 & 36.0 & 33.0 & 39.0 \\ 
College Mathematics & STEM & 34.5 & 33.0 & 35.0 & 29.0                 & \uline{\textbf{31.0}} & 25.0 & 21.0 & 30.0 \\ 
 College Medicine & Other & \textbf{48.0} & 30.6 & 26.0 & 23.1         & 26.0 & 27.8 & 26.6 & 22.0 \\ 
College Physics & STEM & \textbf{28.0} & 26.5 & 24.5 & 24.5            & 21.6 & 22.6 & 24.5 & 14.7 \\ 
 Computer Security & STEM & \textbf{57.0} & 45.0 & 24.0 & 30.0         & 19.0 & 27.0 & 28.0 & 34.0 \\ 
Conceptual Physics & STEM & 36.5 & \textbf{36.6} & 27.7 & 30.2         & 22.1 & 28.5 & 23.8 & 28.9 \\ 
 Econometrics & Social Science & \textbf{33.0} & 23.7 & 24.6 & 25.4    & 30.7 & 23.7 & 24.6 & 21.9 \\ 
Electrical Engineering & STEM & \textbf{50.0} & 26.9 & 29.0 & 24.1     & 26.2 & 29.0 & 23.5 & 28.3 \\ 
 Elementary Mathematics & STEM & \textbf{30.0} & 24.3 & 26.2 & 25.9    & 27.5 & 25.1 & 25.9 & 24.9 \\ 
Formal Logic & Humanities & 29.0 & 27.0 &\uline{\textbf{35.7}}& 24.6   & 20.6 & 16.7 & 15.9 & 28.6 \\ 
 Global Facts & Other & \textbf{37.0} & 29.0 & 30.0 & 31.0             & 30.0 & \uline{\textbf{37.0}} & 33.0 & 23.0 \\ 
High School Biology & STEM & \textbf{48.0} & 34.5 & 25.8 & 26.5        & 25.5 & 24.8 & 24.8 & 29.0 \\ 
  High School Chemistry & STEM & \textbf{33.0} & 28.1 & 27.6 & 19.7                                 & 27.1 & 27.1 & 24.1 & 16.7 \\ 
High School Computer Science& STEM &\textbf{39.0}& 31.0 & 29.0 & 26.0                               & 26.0 & 27.0 & 25.0 & 28.0 \\ 
  High School European History & Humanities & \textbf{54.0} & 44.2 & 23.6 & 28.5                    & 24.9 & 26.7 & 25.5 & 26.7 \\ 
High School Geography & Social Science & \textbf{58.0} & 34.3 & 34.3 & 20.7                         & 19.2 & 17.7 & 22.2 & 21.2 \\ 
  High School Government And Politics & Social Science & \textbf{58.0} & 44.6 & 35.2 & 16.6         & 25.9 & 21.8 & 21.8 & 32.6 \\ 
High School Macroeconomics & Social Science & \textbf{40.5} & 35.4 & 34.4 & 25.9                    & 22.8 & 24.6 & 23.8 & 36.2 \\ 
 High School Mathematics & STEM & 28.0 & 24.8 & 26.7 & 25.2                                         & 28.5 & 26.7 & 25.2 & \uline{\textbf{29.6}} \\ 
High School Microeconomics & Social Science & \textbf{42.0} & 31.9 & 23.5 & 23.1                    & 25.2 & 21.4 & 25.2 & 30.7 \\ 
 High School Physics & STEM & 28.0 & 26.5 & 27.8 & 26.5                                             & 27.2 & \uline{\textbf{29.8}} & 21.9 & 25.8 \\ 
High School Psychology & Social Science & \textbf{61.0} & 47.3 & 32.3 & 23.1                        & 22.9 & 23.7 & 23.8 & 22.9 \\ 
 High School Statistics & STEM & 30.5 & 35.2 & 21.3 & 21.3                                          & 22.2 & 23.2 & 19.9 & \uline{\textbf{45.4}} \\ 
High School Us History & Humanities & \textbf{53.0} & 39.7 & 24.5 & 21.6                            & 24.5 & 27.5 & 24.5 & 25.5 \\ 
 High School World History & Humanities & \textbf{56.0} & 40.9 & 29.1 & 25.7                        & 27.4 & 25.7 & 24.5 & 24.9 \\ 
Human Aging & Other & \textbf{50.0} & 40.8 & 14.8 & 30.5                                            & 30.5 & 27.4 & 37.2 & 24.7 \\ 
  Human Sexuality & Social Science & \textbf{54.0} & 36.6 & 28.2 & 22.1                             & 22.1 & 25.2 & 22.9 & 22.1 \\ 
International Law & Humanities & \textbf{55.5} & 51.2 & 26.5 & 30.6                                 & 32.2 & 30.6 & 39.7 & 26.4 \\ 
  Jurisprudence & Humanities & \textbf{55.0} & 38.9 & 26.9 & 22.2                                   & 27.8 & 25.0 & 26.9 & 24.1 \\ 
Logical Fallacies & Humanities & \textbf{48.0} & 39.3 & 19.6 & 27.0                                 & 23.9 & 27.6 & 29.5 & 24.5 \\ 
  Machine Learning & STEM & 31.0 & 23.2 & 17.9 & \uline{\textbf{33.0}}                              & 28.6 & 30.4 & 23.2 & 25.9 \\ 
Management & Other & \textbf{56.0} & 35.0 & 26.2 & 29.1                                             & 21.4 & 23.3 & 27.2 & 22.3 \\ 
  Marketing & Other & \textbf{60.0} & 46.6 & 22.2 & 24.4                                            & 25.2 & 28.2 & 23.9 & 27.4 \\ 
Medical Genetics & Other & 40.0 & \textbf{43.0} & 27.0 & 24.0                                       & 22.0 & 23.0 & 24.0 & 33.0 \\ 
  Miscellaneous & Other & \textbf{60.0} & 42.4 & 22.5 & 27.5                                        & 29.3 & 26.2 & 27.6 & 24.6 \\ 
Moral Disputes & Humanities & \textbf{44.5} & 40.2 & 29.5 & 25.7                                    & 24.9 & 24.0 & 24.9 & 26.9 \\ 
  Moral Scenarios & Humanities & 26.0 & 24.3 & \uline{\textbf{27.3}} & 24.6                         & 23.8 & 24.6 & 24.3 & 24.2 \\ 
Nutrition & Other & \textbf{47.0} & 37.6 & 28.1 & 23.2                                              & 25.8 & 25.8 & 25.2 & 25.5 \\ 
  Philosophy & Humanities & \textbf{51.0} & 39.9 & 28.0 & 28.9                                      & 29.3 & 28.3 & 26.7 & 30.2 \\ 
Prehistory & Humanities & \textbf{53.0} & 36.1 & 26.5 & 25.9                                        & 26.9 & 27.5 & 29.3 & 27.2 \\ 
  Professional Accounting & Other & \textbf{33.0} & 25.9 & 27.0 & 29.1                              & 27.3 & 27.0 & 27.0 & 22.3 \\ 
Professional Law & Humanities & \textbf{34.5} & 30.2 & 27.1 & 25.0                                  & 24.6 & 26.9 & 25.8 & 26.9 \\ 
  Professional Medicine & Other & 36.0 & \textbf{44.5} & 19.9 & 31.6                                & 21.0 & 27.9 & 22.8 & 44.5 \\ 
Professional Psychology & Social Science & \textbf{44.5} & 35.1 & 26.3 & 27.3                       & 25.2 & 27.5 & 25.5 & 25.8 \\ 
  Public Relations & Social Science & \textbf{48.0} & 40.9 & 33.6 & 30.9                            & 29.1 & 26.4 & 28.2 & 20.0 \\ 
Security Studies & Social Science & \textbf{52.0} & 31.8 & 39.2 & 17.5                              & 21.2 & 16.3 & 18.8 & 37.6 \\ 
  Sociology & Social Science & \textbf{53.0} & 46.8 & 25.4 & 24.4                                   & 24.9 & 23.9 & 22.9 & 21.4 \\ 
Us Foreign Policy & Social Science & \textbf{69.0} & 46.0 & 27.0 & 31.0                             & 25.0 & 28.0 & 24.0 & 23.0 \\ 
  Virology & Other & \textbf{46.0} & 30.1 & 21.7 & 30.1                                             & 27.1 & 28.3 & 31.3 & 29.5 \\ 
World Religions & Humanities & \textbf{55.0} & 50.9 & 27.5 & 25.2                                   & 29.8 & 32.2 & 32.8 & 35.1 \\ 
\midrule
  Humanities &  & \textbf{40.6} & 34.0 & 27.1 & 25.8                                                & 26.2 & 26.4 & 26.9 & 27.0 \\ 
 STEM &  & \textbf{36.7} & 30.5 & 26.5 & 25.8                                                       & 26.1 & 27.1 & 24.4 & 27.3 \\ 
  Social Science &  & \textbf{50.5} & 38.3 & 30.3 & 24.0                                            & 24.5 & 23.3 & 23.6 & 26.8 \\ 
 Other &  & \textbf{49.0} & 38.1 & 24.6 & 27.1                                                      & 25.9 & 26.5 & 27.8 & 26.8 \\ 

\midrule
  All &  & \textbf{43.9} & 35.1 & 27.0 & 25.7                                                       & 25.7 & 26.0 & 25.6 & 26.9 \\ 

\bottomrule
  \end{tabular}}
  \caption{
    \textbf{MMLU.} 5-shot results per domain on the test sets.
    \label{tab:mmluapp}
  }
\end{table*}

\clearpage

\section{Application: Large Batch-size Training on 7B}
\label{sec:7B}

\subsection{7B Training Data Combination}

Our 7B large batch size (LBS) training dataset is primarily based on Slimpajama, however, to obtain a sufficient proportion of web text, we have incorporated additional web data from the Commoncrawl corpus in RedPajama. We have also adjusted the proportions of various data sources in line with our 1.3B model training. For instance, we elevate the sampling frequency of Github and Wikipedia and increase the diversity of data sources by adding S2orc~\cite{lo2019s2orc} and Stack-Markdown~\cite{Kocetkov2022TheStack} following~\cite{mosaicml23}, as detailed in Table~\ref{tab:7B_training_data}. It's crucial to understand that our primary focus is not solely on achieving the best performance. Instead, we place a higher emphasis on optimizing data combinations and ensuring the convergence of training large language models with large batch sizes. Consequently, we continue to utilize the SlimPajama/RedPajama Commoncrawl instead of higher-quality RefinedWeb.

\begin{table}[h]
\centering
\resizebox{0.5\textwidth}{!}{
\begin{tabular}{l|l}
  dataset & proportion \\ \hline 
Slimpj.Arxiv &              4\% (54B) \\
 Slimpj.StackExchanges &     3.2\% (43B)  \\
Slimpj.Github &             4.9\% (66B)  \\
 Slimpj.Wikipedia &          7.5\% (101B)  \\
Slimpj.Books &              4.3\% (57B)  \\
 Slimpj.C4 &                 17.6\% (236B) \\
S2orc &              3\% (40B)  \\
 Markdown &           3\% (40B)  \\
Slimpj.CC &    34.5\% (462B)\\
 Redpaj.CC (ext.) & 18\% (241B) \\ \hline
Total & 1.34T \\
\end{tabular}
}
\vspace{-0.08in}
\caption{Data combination of 7B model training in large batch size style.}
\label{tab:7B_training_data}
\vspace{-0.05in}
\end{table}

\subsection{7B Model Training Configurations}

\begin{table}[h]
\centering
\resizebox{1.0\textwidth}{!}{
\begin{tabular}{l|c|c|c|c|c|c|c}
 Model & n\_params & n\_layers & d\_model & n\_heads & d\_heads & batch size & learning rate \\ \hline
GPT-3 & 6.7B & 32 & 4,096 & 32 & 128 & 2M & 1.2$\times$10-4 \\
 LLaMA & 6.7B & 32 & 4,096 & 32 & 128 & 4M & 3.0$\times$10-4 \\
\bf Our LBS  & 6.7B  &   32  &  4,096  & 32 & 128  &  \bf 14.3M  & 1.8$\times$10-4 \\ 
\end{tabular}
}
\caption{Detailed model sizes, architectures, and optimization hyper-parameters.}
\label{tab:detailed_arch}
\end{table}
\noindent{\bf Architecture.}  For the LBS model training, we adopt MPT architecture~\cite{mosaicml23}, the max sequence length is 2,048. We use Triton~\cite{tillet2019triton} with Flash Attention~\cite{dao2022flashattention} as the self-attention implementation. Alibi is enabled to make model more flexible for input length extrapolation. The total number of parameters is 6.7B.

\noindent{\bf Tokenizer.} The tokenizer used for 7B training is adapted  GPT-NeoX-20b. Following~\cite{mosaicml23}, the model's vocabulary size is adjusted to 50,432 for improved mfu and leaving a few tokens available that can be used in subsequent training.

\noindent{\bf Optimizer.} We employ the AdamW optimizer to train our models, adopting these specific hyper-parameters: $\beta_1$ set at 0.9 and $\beta_2$ at 0.95. We adopt a learning rate schedule that traces a cosine pattern, concluding with a learning rate that is 10\% of its maximum value. Along with this, we use a multi-stage weight decay scheduler as described in Section~\ref{PTWD}, cap the gradient with a clipping value of 1.0, and use a warmup spanning 2,000 steps.

\noindent{\bf System and platform.} For our 7B model training with a large batch size, we use 232 NVIDIA A100 GPUs (80G). We employ llm-foundry~\cite{Foundry} as the training platform. We use FSDP with activation checkpointing enabled to save memory consumption. We also use automatic mixed precision of bf16 in training.

\subsection{Fast Training with Large Batch-size}

Large batch training allows a larger learning rate, leading to a faster convergence of large models. Also, utilizing a larger batch size can optimize hardware resource usage to make training procedures more efficient. Additionally, fewer batches are required, which further accelerates the training process. As shown in Table~\ref{tab:speed}, our large batch training scheme achieves much higher throughput and mfu than LLaMA~\cite{touvron2023llama} and MPT~\cite{mosaicml23} with fewer total training GPU hours. Overall, in a convex optimization framework, leveraging a larger portion of the dataset typically leads to enhanced results. However, for most large deep models that involve non-convex optimizations, the precise nature of the loss landscape remains elusive, making the scenario more intricate.
Many prior works~\cite{hoffer2017train,keskar2016large} have noticed that training with larger batches often results in overfitting compared to those using smaller batch sizes for the same network. When utilizing large batch training, there is a propensity for the model to become stuck or even gravitate towards potential saddle points within the loss landscape. While large batch training methods often focus on the nearest relative minima they encounter, networks trained with smaller batches usually navigate the loss landscape more thoroughly before committing to an optimal minimum. The minima reached through large batch training can be distinctly different from those achieved with smaller batch training methods. In the following, we introduce an approach to mitigate overfitting when training large language models in a large batch-size scheme.

\begin{table}[h]
\centering
\resizebox{0.99\textwidth}{!}{
\begin{tabular}{l|c|c|c|c|c}
 model&batch size &  \# GPUs (A100-80G)  & throughput  & mfu &  GPU-hours   \\  \hline
LLaMA-7B &4M  &  -- & --      & --  & 82,432 \\
 MPT-7B&4M  &  232  & 3,310        & 0.4575 &  84.351  \\ 
LBS-7B (ours)&\bf 14M &  232  & \bf 3,626        & \bf 0.5011  & \bf 76,999 \\ 
\end{tabular}
}
\caption{Training speed of throughput (tokens per sec on each GPU), {\em model FLOPs utilization} (mfu)~\cite{chowdhery2022palm} and total GPU-hours (per trillion training tokens).}
\label{tab:speed}
\end{table}

\subsection{Progressive Training on Weight Decay} \label{PTWD}

\begin{figure*}[h]
\vspace{-0.1in}
\begin{center}
\includegraphics[width=0.9\textwidth]{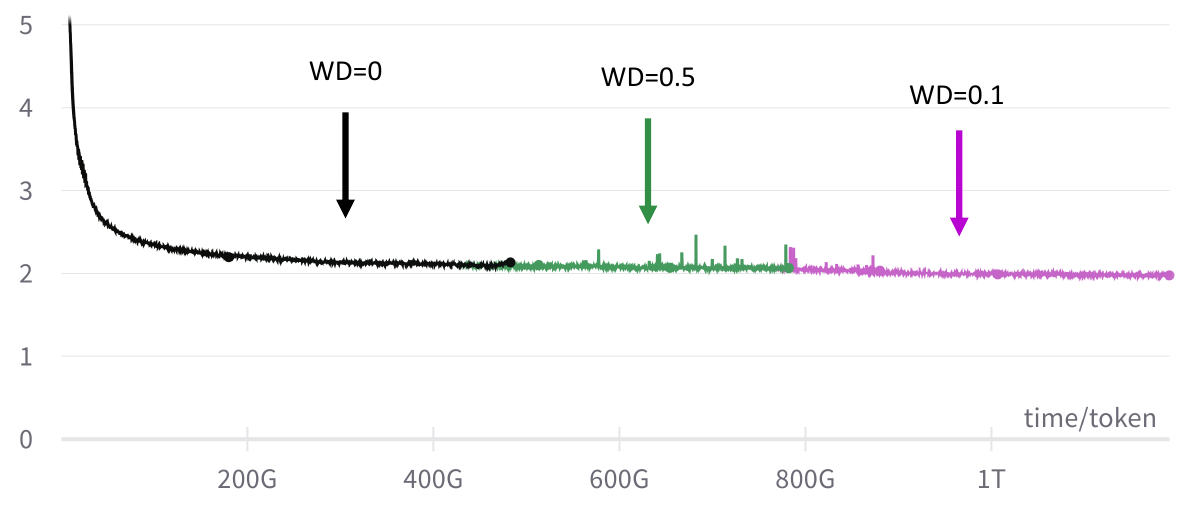}
\end{center}
\vspace{-0.1in}
\caption{Loss curve of our LBS-7B training.}
\label{fig:7B_loss}
\vspace{-0.03in}
\end{figure*}

Prior work~\cite{liu2023dropout} observed that dropout operation is utilized only in the early stages of training and is deactivated in subsequent phases. Models that incorporate this early dropout strategy tend to exhibit reduced final training loss compared to models that do not use dropout. In contrast to this, our approach emphasizes the role of weight decay during large model training. We introduce a novel training strategy for large language models, wherein the training process is segmented into various stages. Within each stage, a distinct weight decay is applied to the model to serve specific objectives. We've termed this approach {\em Progressive Training on Weight Decay} (PTWD). Owing to this methodology, our model, even when trained with a large batch size and extremely small iterations, achieves smooth convergence. As illustrated in Fig.~\ref{fig:7B_loss}, our training strategy consists of three distinct phases. Initially, we negate weight decay by setting it to zero and allow the model to train until full convergence is achieved. It usually can reach a lower loss level within this stage compared to using weight decay, even if it slightly overfits. Following this, in the second phase, we introduce a substantial weight decay, with a value of 0.5 in our experiments, to suppress the overfitting. Once the loss values stabilize, we transition to the third phase, wherein a standard weight decay of 0.1 is implemented, a value consistent with many other LLMs training. Intriguing, each phase spontaneously converges to roughly 1/3 of the total training budget, ensuring effective allocation of training budget throughout the process.

\subsection{Results of Pre-training and Instruction Tuning} 

The results from our pretraining and subsequent instruction tuning on ShareGPT dataset are presented in Table~\ref{tab:7B_results}. Notably, after instruction tuning, there is a significant enhancement in MMLU and TruthfulQA metrics. In contrast, the performance on ARC and HellaSwag has a slight decrease. On the whole, the average accuracy has a substantial boost following instruction tuning. More evaluation results on the pretrained LBS model are provided in Table~\ref{tab:six_conf_more_res}. 

\begin{table}[h]
\centering
\resizebox{0.99\textwidth}{!}{
\begin{tabular}{l|c|cccc}
 \textbf{Model} & \textbf{Average} & \textbf{ARC} & \textbf{HellaSwag} & \textbf{MMLU} & \textbf{TruthfulQA} \\ \hline
Ours-LBS-7B-Base  &  44.1	& 44.3	& 69.8	& 26.1	& 36.1       \\  
 Ours-LBS-7B-Instruct  & 46.4 & 43.5 &	68.0 &	32.1 &	42.1 \\
\end{tabular}
}
\caption{Results of our large batch-size (LBS) trained 7B models following Huggingface Leaderboard Evaluation~\cite{open-llm-leaderboard} using Harness~\cite{eval-harness}.}
\label{tab:7B_results}
\end{table}

\end{document}